\UseRawInputEncoding

\documentclass[conference]{IEEEtran}
\usepackage{cite}
\usepackage{amsmath,amssymb,amsfonts}
\usepackage{algorithmic}
\usepackage{graphicx}
\usepackage{hyperref}
\usepackage{subcaption}
\usepackage{booktabs}
\usepackage{multirow}
\usepackage[utf8]{inputenc}
\usepackage{textcomp}
\usepackage{makecell}

\usepackage{xcolor}
\def\BibTeX{{\rm B\kern-.05em{\sc i\kern-.025em b}\kern-.08em
    T\kern-.1667em\lower.7ex\hbox{E}\kern-.125emX}}
\begin{document}

\title{Lightweight image segmentation for echocardiography}

\author{
\IEEEauthorblockN{
Anders Kjelsrud\IEEEauthorrefmark{1},
Lasse Løvstakken\IEEEauthorrefmark{2},
Erik Smistad\IEEEauthorrefmark{2}\IEEEauthorrefmark{3},
Håvard Dalen\IEEEauthorrefmark{2}\IEEEauthorrefmark{4},
and Gilles Van De Vyver\IEEEauthorrefmark{2}}
\IEEEauthorblockA{\IEEEauthorrefmark{1}\textit{Intility}, Oslo, Norway}
\IEEEauthorblockA{\IEEEauthorrefmark{2}\textit{Norwegian University of Science and Technology}, Trondheim, Norway}
\IEEEauthorblockA{\IEEEauthorrefmark{3}\textit{SINTEF Medical Image Analysis}, Trondheim, Norway}
\IEEEauthorblockA{\IEEEauthorrefmark{4}\textit{St. Olav's University Hospital}, Trondheim, Norway}
}

\maketitle

\begin{abstract}
Accurate segmentation of the left ventricle in echocardiography can enable fully automatic extraction of clinical measurements such as volumes and ejection fraction. While models configured by nnU-Net perform well, they are large and slow, thus limiting real-time use.

We identified the most effective components of nnU-Net for cardiac segmentation through an ablation study, incrementally evaluating data augmentation schemes, architectural modifications, loss functions, and post-processing techniques. Our analysis revealed that simple affine augmentations and deep supervision drive performance, while complex augmentations and large model capacity offer diminishing returns.

Based on these insights, we developed a lightweight U-Net (2M vs 33M parameters) that achieves statistically equivalent performance to nnU-Net on CAMUS (N=500) with Dice scores of 0.93/0.85/0.89 vs 0.93/0.86/0.89 for LV/MYO/LA ($p>0.05$), while being 16 times smaller and 4 times faster (1.35ms vs 5.40ms per frame) than the default nnU-Net configuration. 
Cross-dataset evaluation on an internal dataset (N=311) confirms comparable generalization.
\end{abstract}

\section{Introduction}
Cardiovascular disease (CVD) is the leading global cause of death, responsible for 20.5 million lives lost in 2021. Analyzing 2D echocardiographic images allows for extraction of clinical measurements such as left ventricular ejection fraction and volume, which assess cardiovascular health. Manual or semi-automatic segmentation is the standard in clinics today. Fully automatic cardiac segmentation could reduce manual workload, free up clinician time for patient care, and provide more robust measurements. 

The nnU-Net framework, introduced in 2018, automatically configures U-Net pipelines without manual tuning and achieved state-of-the-art performance across various datasets \cite{isensee2018nnunet}. New architectures have been adapted to the medical imaging domain, and many claim to outperform CNN-based U-Nets.
However, the creators of nnU-Net show that CNN-based U-Nets still give the best performance for most medical imaging datasets if tuned well, and advocate that future research should prioritize thorough validation over introducing novel architectures \cite{new_nnunet}.

This work conducts an ablation study to identify the key performance-driving components of nnU-Net. To this end, we break down nnU-Net and run a series of experiments where various components are disabled. Then, the components are added to and evaluated on a lightweight U-Net. A smaller model is desirable because it trains faster, offers better generalization to new datasets, and has faster inference time which is particularly useful when deployed on portable ultrasound scanners with limited compute for real-time inference.
\section{Method}
To bridge the performance gap between nnU-Net and a lightweight U-Net on the CAMUS dataset, we incrementally add components from nnU-Net to a smaller U-Net. Successful additions showing a performance increase are retained for downstream experiments. Models are evaluated using a per-class Dice coefficient and pixel-wise Hausdorff distance, and a count of anatomical outliers. Statistical significance is tested with the Wilcoxon signed-rank test. 

\subsection{Datasets}
\subsubsection{CAMUS}
The CAMUS dataset \cite{camus-dataset} consists of 2D echocardiographic sequences from 500 patients, with annotations for the left ventricle (LV), myocardium (MYO), and left atrium (LA). It includes 2000 image-annotation pairs, each $256 \times 256$ pixels, captured using GE Vivid E95 scanners.

\subsubsection{HUNT4}
The HUNT4Echo dataset \cite{hunt4} originates from a large-scale health study in Trøndelag, including 2462 LV-focused echocardiographic exams. Among these, 311 exams provide segmentation labels for LV, MYO, and LA.

\subsubsection{Differences}
\label{sec:dataset_diffs}
\begin{figure}[htbp]
  \centering
    \includegraphics[width=\linewidth]{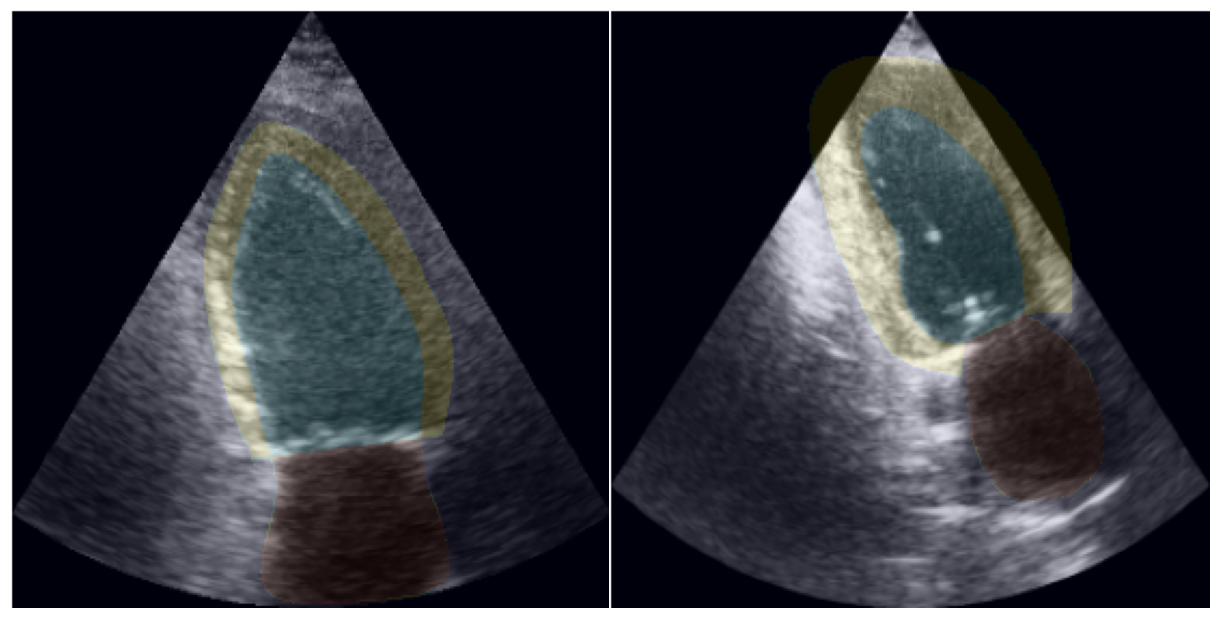}
  \caption{Ground truth segmentations from HUNT4 (left) and CAMUS (right). Images from HUNT4 are consistently LV-focused, while those from CAMUS are not.}
  \label{fig:camus_hunt4}
\end{figure}

Although both HUNT4 and CAMUS present the same clinical views, the two datasets cannot be directly combined and jointly trained on due to differences in annotation conventions and imaging techniques. In CAMUS, the myocardium is annotated significantly thicker than in HUNT4, as illustrated in Figure \ref{fig:camus_hunt4}. Additionally, HUNT4 images have less variance in terms of positioning and are regarded as more standardized \cite{vandevyver2023robust}.

\subsubsection{Dataset usage and splits}
CAMUS is used for development, testing, and tuning, while HUNT4 is used to evaluate the models ability to generalize across different imaging conditions. An $80/10/10$ train/validation/test split is applied to both datasets, yielding 1600 training, 200 validation, and 200 test images for CAMUS, and 846 training, 105 validation, and 105 test images for HUNT4.

\subsection{Baselines}
Two baselines are established. One using the U-Net 1 architecture from the CAMUS study \cite{camus-dataset}, and another for nnU-Net \cite{isensee2018nnunet}, see Table \ref{table:baselines} for a comparison. The goal of this work is to analyze the nnU-Net by incrementally adding its components to the U-Net 1 baseline in order to find the most important parts in terms of performance. All experiments using the lightweight U-Net use the following training scheme:
\begin{itemize}
    \item $100$ epochs with early stopping (patience=$50$)
    \item Adam optimizer with initial learning rate of $0.001$ 
    \item Batch size $32$
\end{itemize}
For a fair comparison, the nnU-Net model is also trained for $100$ epochs. This should be unproblematic as empirical observations show convergence well before this point.

\begin{table*}[ht]
\centering
\footnotesize
\
\begin{tabular}{l l l l l l l l}
\toprule
\textbf{Model} & \textbf{\makecell[l]{Number of \\ channels}} & \textbf{\makecell[l]{Upsampling \\ scheme}} & \textbf{\makecell[l]{Downsampling \\ scheme}} & \textbf{\makecell[l]{Normalization \\ scheme}} & \textbf{\makecell[l]{Loss \\ function}} & \textbf{\makecell[l]{ Trainable \\ Params}} & \textbf{Augmentations} \\ \midrule
U-Net 1 & 32 $\downarrow$ 128 $\uparrow$ 16 & \makecell[l]{Nearest neighbor \\ interpolation} & 2$\times$2 Maxpool & None & Dice & 2M & None \\ \\
nnU-Net & 32 $\downarrow$ 512 $\uparrow$ 32 & \makecell[l]{Transposed \\ convolutions} & \makecell[l]{Strided \\ convolutions} & InstanceNorm & Dice+CE & 33M & \makecell[l]{Rotation, scaling, Gaussian noise/blur,\\ brightness, contrast, low-res simulation, \\ gamma correction, mirroring}
 \\\bottomrule
\end{tabular}
\caption{Architectural differences between the two baselines. Number of channels denotes the channel counts at the first, deepest, and final feature maps. The minimum feature map resolutions are $8\times 8$ for U-Net 1 and $4\times 4$ for nnU-Net.}
\label{table:baselines}
\end{table*}

\subsection{Experiments} \label{method:experiments}
Several components of nnU-Net are systematically analyzed through incremental experiments:
\begin{enumerate}
    \item Normalization scheme: Instance normalization (IN) and Batch normalization (BN).
     
     \item Activation functions: ReLU, Leaky ReLU, Mish, and GELU.
     
     \item During post-processing, connected component analysis (CCA) and the morphological operations opening and closing are used to exploit the fact that the images only contain one instance of each target structure.

     \item Data augmentation following Table \ref{table:aug_configs}.

     \item Loss functions: Dice and the average of Dice and Cross Entropy.

    \item Deep Supervision (DS) uses
    $1\times 1$ convolutions to reduce the number of channels down to the number of segmentation classes. Then, the feature maps are upscaled using nearest neighbor interpolation, and finally softmaxed. A new loss function is chosen to accommodate DS: 
        $L = L_{\text{final}} + \sum^n_{i=1} \lambda_i L_{\text{auxiliary}_i}$,
    where $\lambda$ is set to $0.3$ for all $i$ , $L_{\text{final}}$ is the loss between the final prediction and the ground truth, $L_{\text{auxiliary}_i}$ is the loss between the upscaled intermediate feature map $i$ and the ground truth, and n the number of deep supervision levels.

    \item Three additional networks with varying sizes, shown in Table \ref{table:sizes}, are trained. The goal is to find the needed capacity of a model to learn the distribution of the dataset.

    \item Architectural changes: Residual connections, and transposed convolutions vs. simple upsampling.
\end{enumerate}

\begin{table}
\centering
\small 
\resizebox{\columnwidth}{!}{%
\begin{tabular}{lcccccccc}
\toprule
\textbf{Config} 
& \textbf{Shift} 
& \textbf{Scale} 
& \textbf{Rot.} 
& \textbf{Gamma} 
& \textbf{Noise} 
& \textbf{Blackout} 
& \textbf{\shortstack{Bri.\\Con.\\Sat.}} 
& \textbf{Hue} \\ \midrule
I   & $(-0.05, 5)$  & $(-0.1, 0.05)$ & $(-5, 5)$     & $(90, 110)$ & $(5, 10)$  & $0.25$ & $(0.95, 1.05)$ & $(-0.05, 0.05)$ \\
II  & $(-0.10, 10)$  & $(-0.2, 0.1)$  & $(-10, 10)$   & $(85, 115)$ & $(10, 25)$ & $0.25$ & $(0.90, 1.10)$ & $(-0.10, 0.10)$ \\
III & $(-0.15, 15)$  & $(-0.3, 0.15)$ & $(-15, 15)$   & $(80, 120)$ & $(20, 35)$ & $0.25$ & $(0.85, 1.15)$ & $(-0.15, 0.15)$ \\
IV  & $(-0.20, 20)$  & $(-0.4, 0.20)$ & $(-20, 20)$   & $(70, 130)$ & $(30, 45)$ & $0.25$ & $(0.80, 1.20)$ & $(-0.20, 0.20)$ \\
V   & $(-0.25, 25)$  & $(-0.5, 0.25)$ & $(-30, 30)$   & $(60, 140)$ & $(40, 50)$ & $0.25$ & $(0.75, 1.25)$ & $(-0.25, 0.25)$ \\ Affine & $(-0.1, 0.1)$ & $(-0.2, 0.1)$ & $(-10, 10)$  & - & - & - & - & - \\ \bottomrule
\end{tabular}%
}
\caption{Data augmentation experiment plan. Blackout is the probability of placing a rectangle with sides $\sim\mathcal{U}(25, 65)$ px randomly. Other parameters are inputs to Albumentations-augmentations.}
\label{table:aug_configs}
\end{table}

\begin{table}
\centering
\small
\resizebox{\columnwidth}{!}{%
\begin{tabular}{lllllll}
\toprule
\textbf{Arch.} 
& \textbf{\shortstack{Channels\\(start$\downarrow$end$\uparrow$final)}} 
& \textbf{\shortstack{Low\\Res.}} 
& \textbf{BS} 
& \textbf{Scheduler} 
& \textbf{\shortstack{\# Params}} 
\\ \midrule
U-Net 1 & $32 \downarrow 128 \uparrow 16$ & $8 \times 8$   & 32 & -                   & 2M   \\
U-Net 2 & $64 \downarrow 512 \uparrow 32$ & $8 \times 8$   & 16 & ReduceLROnPlateau   & 11M  \\
U-Net 3 & $64 \downarrow 1024\uparrow 64$ & $16 \times 16$ & 8  & ReduceLROnPlateau   & 31M  \\
U-Net 0 & $16 \downarrow 64 \uparrow 16$  & $64 \times 64$ & 64 & ReduceLROnPlateau   & 119K \\ \bottomrule
\end{tabular}%
}
\caption{Architectural configurations used to evaluate the effect of network size. Normalization scheme, loss function, and augmentations are consistent across models.}

\label{table:sizes}
\end{table}
\section{Results}

\subsection{Baselines}
The two baseline models are trained following the configuration in Table \ref{table:baselines}, with metrics as shown in Table \ref{table:baselines_results}. The table shows the baseline nnU-Net outperforms the U-Net baseline.

\begin{table}[ht]
\centering
\begin{tabular}{l|ll}
\toprule
\textbf{Metric} & \textbf{U-Net 1} & \textbf{nnU-Net} \\ \midrule
Dice LV                        & 0.90   & 0.93     \\
Dice MYO                       & 0.82   & 0.86     \\
Dice LA                        & 0.86   & 0.89     \\
Hausdorff LV                   & 10.13  & 7.10     \\
Hausdorff MYO                   & 12.87  & 9.21     \\
Hausdorff LA                 & 12.04  & 9.31     \\
\# Anatomical Outliers         & 42     & 6     \\   
\bottomrule
\end{tabular}
\caption{Baseline metrics.}
\label{table:baselines_results}
\end{table}

\subsection{Experiments on nnU-Net}
Disabling data augmentation from the nnU-Net baseline results in the Dice scores $0.92$, $0.85$, $0.88$ and Hausdorff distances $10.01$, $10.36$, $10.92$. The difference in both metrics compared to the baseline are statistically significant with $p < 0.05$, using the Wilcoxon signed-rank test, except the Dice score for the LA class. 
With deep supervision enabled, the metrics show no statistically significant difference from the model with only data augmentations disabled, indicating that for nnU-Net, augmentations are more important than deep supervision.

Next, the connected component analysis done by nnU-Net's postprocessing is investigated. For the baseline, postprocessing does not improve performance significantly. For the model trained without augmentations, both metrics improve for LV and MYO, but not for LA. The model trained without augmentations or deep supervision is statistically significantly worse for all classes on both metrics with $p < 0.05$ when postprocessing is enabled. The largest discrepancy is in the Hausdorff distance, because it measures the maximum distance, and is thus more prone to outliers, which the postprocessing reduces.

\subsection{Creating an accurate lightweight U-Net}

Starting from the baseline U-Net 1, Table \ref{table:unet1-experiments} summarizes the results for the top-performing variation for each addition outlined in \ref{method:experiments}. Components showing performance improvements are carried forward for subsequent runs. Next, we show detailed results from experiments on select components.

\subsubsection*{Data augmentation}
Table \ref{table:da_results} shows the results from running experiments with the configurations in Table \ref{table:aug_configs}.
They show that, out of the tested configurations, a moderate augmentation scheme, one applying only affine transformations, is optimal. 
It results in a significantly higher Dice score and lower Hausdorff distance across all classes ($p < 0.05$).
In contrast, the nnU-net-like augmentations (Configs I-V) lead to the lowest number of outliers, but achieve worse Dice scores and Hausdorff distances.

\begin{table}[ht]
\centering
\resizebox{\columnwidth}{!}{%
\begin{tabular}{l|ccccccc}
\toprule
\textbf{Config}
& \shortstack{Dice\\LV} 
& \shortstack{Dice\\MYO} 
& \shortstack{Dice\\LA} 
& \shortstack{Haus.\\LV} 
& \shortstack{Haus.\\MYO}
& \shortstack{Haus.\\LA}
& \shortstack{\# Anatomical\\Outliers} \\
\midrule
Baseline                & 0.90 & 0.82 & 0.86 & 10.13 & 12.87 & 12.04 & 42 \\
Baseline w/ BN          & 0.92 & 0.84 & 0.87 & 8.40  & 11.21 & 10.47 & 26 \\
I                       & 0.92 & \textbf{0.85} & \textbf{0.88} & 8.38  & 11.13 & 9.93  & 12 \\
II                      & 0.92 & \textbf{0.85} & \textbf{0.88} & 8.22  & 10.96 & 10.56 & \textbf{10} \\
III                     & 0.92 & 0.84 & 0.87 & 8.57  & 11.13 & 10.72 & \textbf{10} \\
IV                      & 0.91 & 0.83 & 0.87 & 9.46  & 12.67 & 10.94 & 14 \\
V                       & 0.82 & 0.58 & 0.69 & 27.27 & 55.12 & 26.17 & 66 \\
Affine                  & \textbf{0.93} & \textbf{0.85} & \textbf{0.88} & \textbf{7.80} & \textbf{10.00} & \textbf{9.91} & 12 \\
\bottomrule
\end{tabular}
}
\caption{Dice score and Hausdorff distance for U-Net 1 with different augmentation configurations (see Table \ref{table:aug_configs})}
\label{table:da_results}
\end{table}

\subsubsection*{Deep supervision}

Introducing deep supervision to the baseline U-Net with batch normalization yields a statistically significant improvement in Hausdorff distance for the LV, while performance for other metrics and labels is worse.
When affine augmentations are combined with DS, both evaluated metrics are significantly improved relative to training without deep supervision.

\begin{figure}[htbp]
  \centering
    \includegraphics[width=\linewidth]{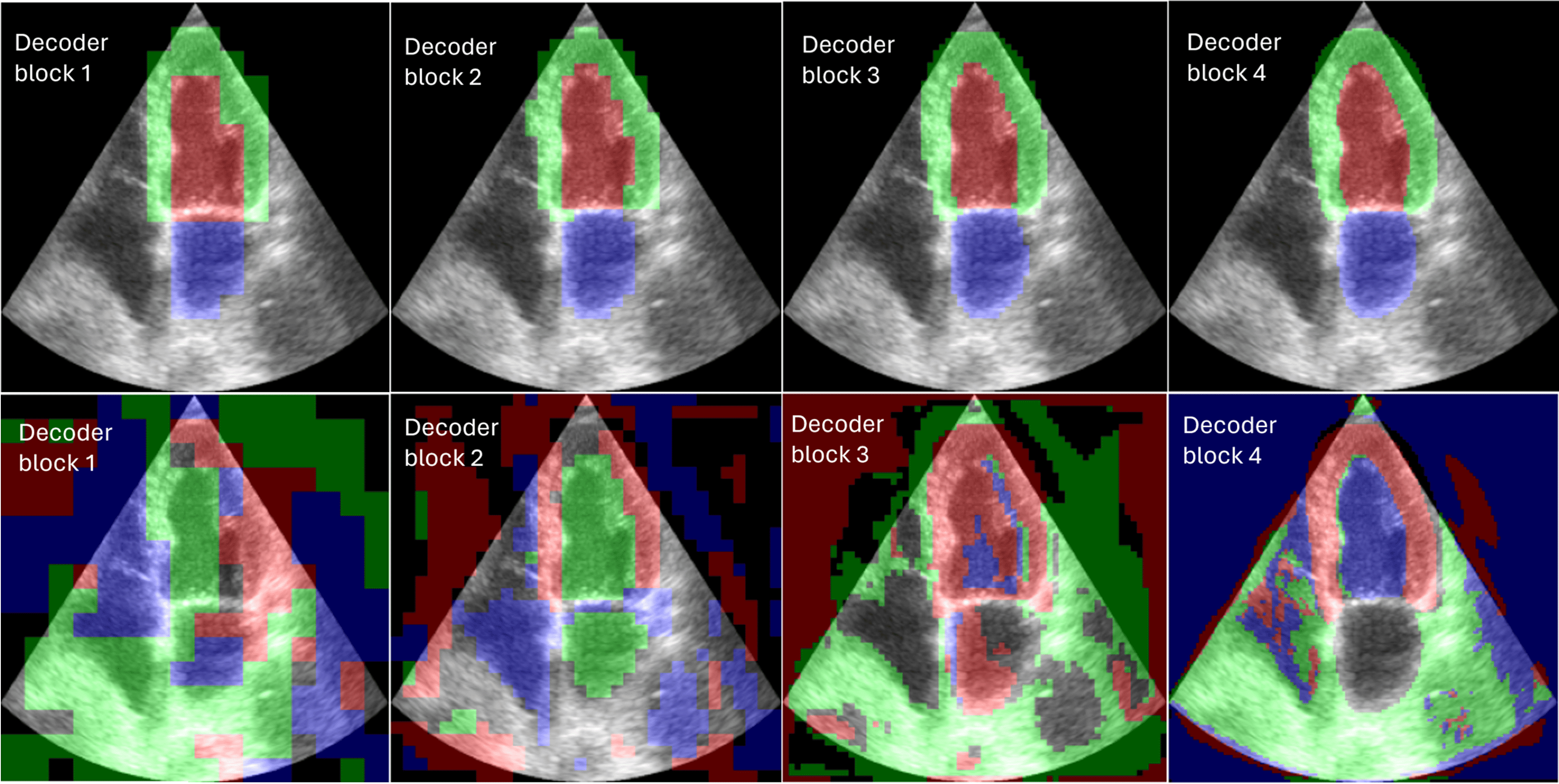}
  \caption{Upscaled intermediate decoder segmentations: top row shows results with deep supervision, bottom row without. Maps from early decoder layers appear pixelated due to an upscaling factor of $16$.}
  \label{fig:ds_comb}
\end{figure}

Upscaled intermediate decoder segmentations from models trained with and without DS are visualized in Figure \ref{fig:ds_comb}. Due to the loss imposed on the intermediate decoder blocks, the model trained with DS produces more anatomically consistent segmentations compared to those generated without.

\begin{table}[ht]
\centering
\small
\resizebox{\columnwidth}{!}{%
\begin{tabular}{l|ccc|ccc|l}
\toprule
\textbf{Addition} 
& \shortstack{Dice\\LV} 
& \shortstack{Dice\\MYO} 
& \shortstack{Dice\\LA} 
& \shortstack{Haus.\\LV} 
& \shortstack{Haus.\\MYO} 
& \shortstack{Haus.\\LA} 
& \textbf{Conclusion} \\ \midrule
Baseline             & 0.90 & 0.82 & 0.86 & 10.13 & 12.87 & 12.04 & - \\
Normalization        & 0.91 & 0.84 & 0.86 & 10.08 & 11.64 & 12.38 & use BN \\
Activation functions & 0.92 & 0.84 & 0.87 & 8.97  & 12.27 & 11.15 & use MISH \\
Postprocessing       & 0.92 & 0.84 & 0.87 & 8.40  & 11.21 & 10.47 & use CCA \\
Data augmentation    & 0.93 & 0.85 & 0.88 & 7.80  & 10.00 & 9.91  & use affine \\
Loss function        & 0.93 & 0.85 & 0.89 & 7.75  & 10.46 & 9.58  & use avg(Dice, CE) \\
Deep supervision     & 0.93 & 0.85 & 0.89 & 7.49  & 9.93  & 9.22  & apply \\
Size                 & 0.93 & 0.85 & 0.88 & 8.22  & 10.52 & 9.54  & discard \\
Architecture         & 0.92 & 0.85 & 0.88 & 8.10  & 10.38 & 9.62  & discard \\ \bottomrule
\end{tabular}%
}
\caption{Summary of metrics for each anatomical structure for experiments on U-Net 1. Conclusion indicates which additions are kept for downstream experiments.}
\label{table:unet1-experiments}
\end{table}

\subsection{The final model}
The final lightweight model extends U-Net 1 by integrating all components showing an increase in segmentation performance. The model is summarized in Table \ref{table:res_final}.
It achieves the Dice scores $0.93, 0.85, 0.89$ and the Hausdorff distances $7.49, 9.93, 9.22$ for the LV, MYO and LA respectively. These results are not statistically different from the results of nnU-Net, shown in table \ref{table:baselines_results}.

\begin{figure}[htbp]
    \centering
    \begin{subfigure}[b]{0.49\columnwidth}
        \includegraphics[width=\linewidth]{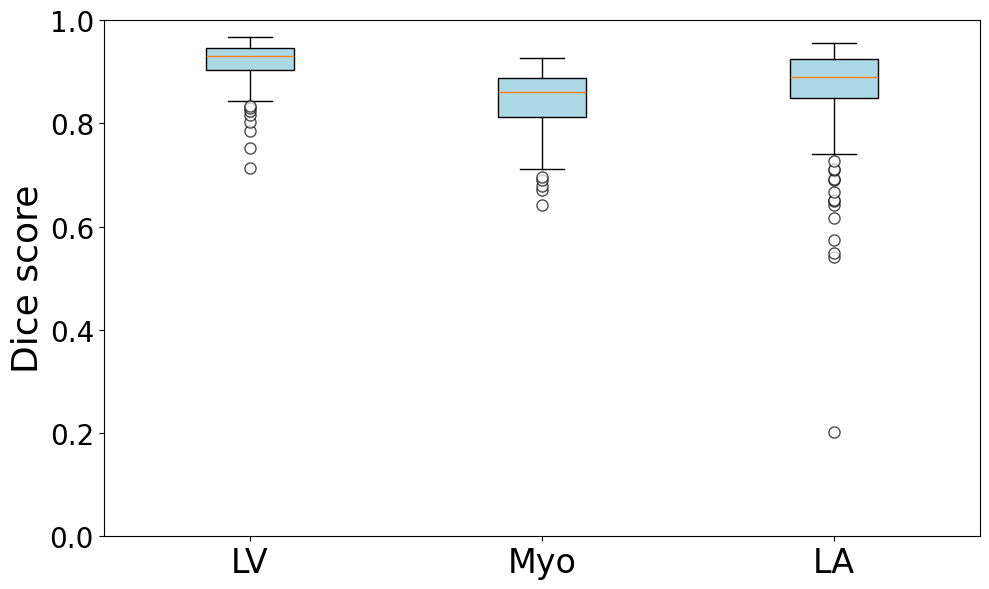}
    \end{subfigure}
    \hfill
    \begin{subfigure}[b]{0.49\columnwidth}
        \includegraphics[width=\linewidth]{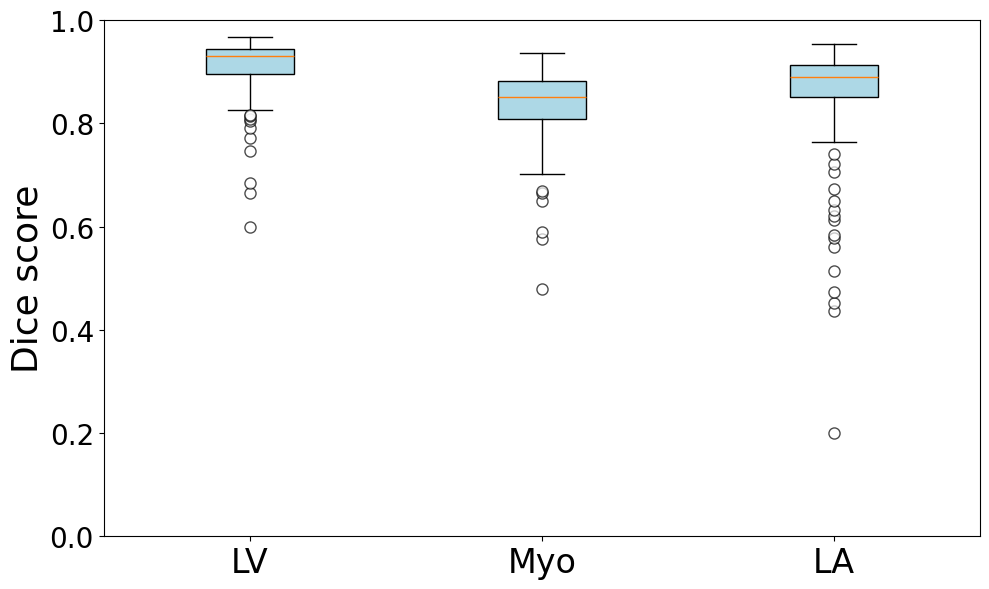}
    \end{subfigure}

    \begin{subfigure}[b]{0.49\columnwidth}
        \includegraphics[width=\linewidth]{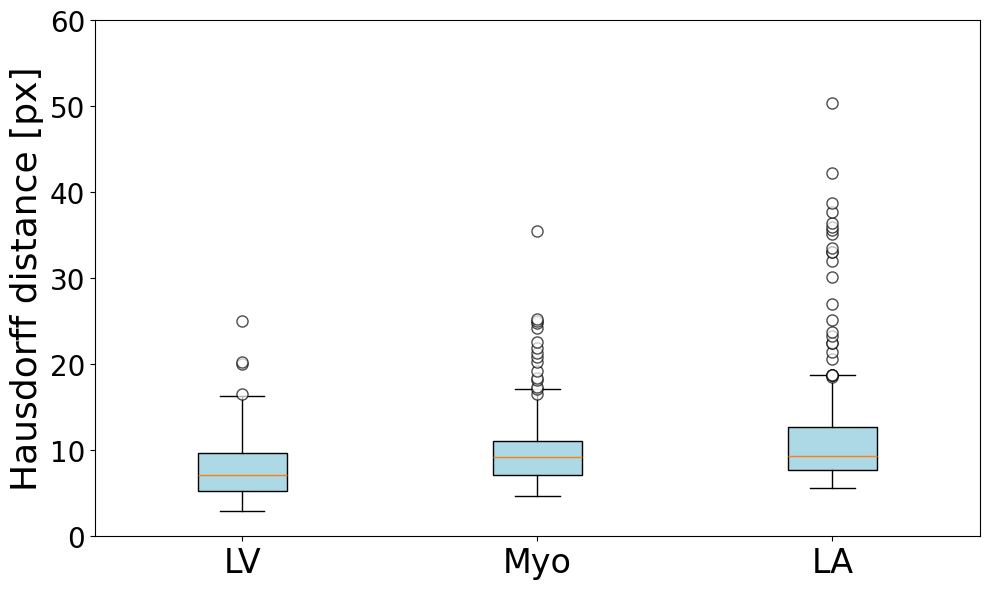}
        \caption{nnU-Net}
    \end{subfigure}
    \hfill
    \begin{subfigure}[b]{0.49\columnwidth}
        \includegraphics[width=\linewidth]{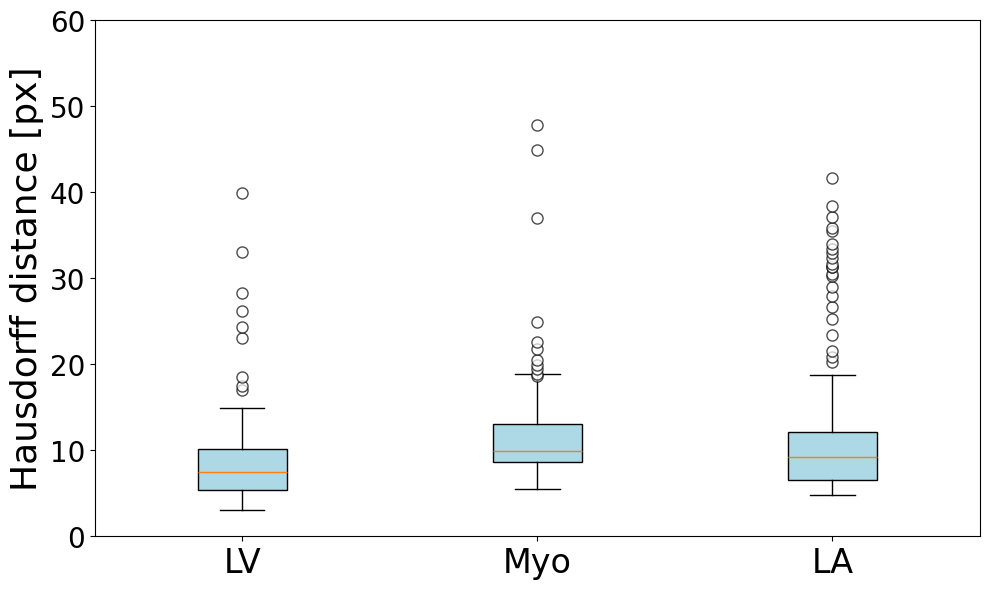}
        \caption{Lightweight U-Net}
    \end{subfigure}
    \caption{Boxplots of Dice scores (top) and Hausdorff distances (bottom) for nnU-Net and our final lightweight U-Net.}
    \label{fig:final_box}
\end{figure}

Figure \ref{fig:final_box} presents boxplots of both metrics for the lightweight U-Net and the nnU-Net baseline. The Wilcoxon signed-rank test shows no statistically significant difference between the two models, with $p=0.87$ for Dice score and $p=0.56$ for Hausdorff distance. 

Compared to nnU-Net, the average inference time per frame was reduced from 5.40 to 1.35 milliseconds when evaluated using PyTorch on an NVIDIA GeForce RTX 3070 Ti Laptop GPU.

\begin{table}[t]
\small 
\centering
\resizebox{\columnwidth}{!}{%
\begin{tabular}{l|l}
\toprule
\textbf{Component} & \textbf{Description} \\ \midrule
Network baseline arch. & U-Net 1 \\
Downsampling & $2\times 2$ Maxpool \\
Upsampling & Nearest neighbor interpolation \\
Normalization & Batch normalization \\
Postprocessing & Connected component analysis \\
Data augmentation & Shift(0.1), Scale(-0.2, 0.1), Rotate(10) \\
Loss function & Dice + Cross Entropy (avg.) \\
Activation function & Mish \\
Deep supervision & Yes, $\lambda=0.3$ for all aux. losses \\
\bottomrule
\end{tabular}
}
\caption{Details of the final lightweight model.}
\label{table:res_final}
\end{table}

\subsection{Results on HUNT4}
Table \ref{table:final_model_camus_hunt} compares the cross-dataset generalization performance of the final lightweight model and the nnU-Net baseline. It reports Dice scores when each model is trained on either CAMUS or HUNT4 and evaluated on both datasets.
The models achieved comparable Dice scores in-domain across all cardiac structures. Cross-dataset evaluation revealed a decrease in Dice scores, with the MYO showing the largest reduction.

\begin{table}[t]
    \centering
    \setlength{\tabcolsep}{4.5pt} 
    \renewcommand{\arraystretch}{1.2}
    \begin{tabular}{llccc|ccc}
        \toprule
        \multirow{2}{*}{\textbf{Trained on}} & 
        \multirow{2}{*}{\textbf{Model}} & 
        \multicolumn{3}{c|}{\textbf{Test: CAMUS}} & 
        \multicolumn{3}{c}{\textbf{Test: HUNT4}} \\
        \cmidrule(lr){3-5} \cmidrule(lr){6-8}
        & & LV & MYO & LA & LV & MYO & LA \\
        \midrule
        CAMUS  & LW U-Net & 0.93 & 0.85 & 0.89 & 0.89 & 0.68 & 0.86 \\
               & nnU-Net & 0.93 & 0.86 & 0.89 & 0.89 & 0.70 & 0.85 \\
        \midrule
        HUNT4  & LW U-Net & 0.86 & 0.66 & 0.75 & 0.95 & 0.82 & 0.91 \\
               & nnU-Net & 0.83 & 0.68 & 0.80 & 0.95 & 0.82 & 0.90 \\
        \bottomrule
    \end{tabular}
    \caption{Dice scores for cross-dataset evaluation of LW U-Net and nnU-Net on CAMUS and HUNT4.}
    \label{table:final_model_camus_hunt}
\end{table}

\section{Discussion}
In this work, components from the nnU-Net pipeline were evaluated and used to design a lightweight U-Net with comparable performance to a full nnU-Net trained model on the CAMUS and HUNT4 datasets. The most influential elements are normalization layers, deep supervision, and data augmentation, with the latter having the biggest impact. While augmented images may lack clinical realism, they still improve performance.

For LV segmentation, the full nnU-Net augmentation scheme is excessive. Simple affine transformations, such as shifting, scaling, and rotation yielded the best results, reducing anatomical outliers from 26 to 12 and improving Dice score and Hausdorff distance. Combining these augmentations with DS further boosts performance, guiding intermediate decoder layers toward anatomically plausible shapes, as shown in Figure \ref{fig:ds_comb}. This aligns with prior findings \cite{ds_unet}, which show that training U-Nets with DS increases segmentation accuracy.

Cross-dataset tests show a performance drop, particularly for MYO, likely due to domain shifts arising from the annotation differences outlined in \ref{sec:dataset_diffs}. The small differences between the lightweight and nnU-Net models suggests dataset variability, rather than model complexity, as the main factor limiting generalization.

The final model is 16 times smaller than nnU-Net, with 2 instead of 33 million parameters, enabling faster inference, lower power consumption, and potential deployment on portable scanners for real-time use. These findings suggest that although nnU-Net's comprehensive design allows it to be applied to many medical imaging tasks, certain complexities are unnecessary for cardiac ultrasound segmentation and can be removed without loss of performance.

\bibliographystyle{ieeetr}  
\bibliography{references}

\end{document}